\documentclass{article}

\usepackage{arxiv}

\usepackage[utf8]{inputenc} 
\usepackage[T1]{fontenc}    
\usepackage{hyperref}       
\usepackage{url}            
\usepackage{booktabs}       
\usepackage{amsfonts}       
\usepackage{nicefrac}       
\usepackage{microtype}      
\usepackage{lipsum}		
\usepackage{graphicx}
\usepackage{doi}

\title{Predicting Lung Disease Severity via Image-Based AQI Analysis using Deep Learning Techniques}

\author{ 
{\includegraphics[scale=0.06]{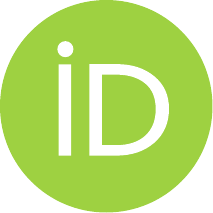}\hspace{1mm}Anvita Mahajan}\thanks{Use footnote for providing further
		information about author (webpage, alternative
		address)---\emph{not} for acknowledging funding agencies.} \\
	Department of Computer Engineering\\
	  COEP Technological University\\
	Pune, India 411005 \\
	\texttt{mahajanaa20.comp@coeptech.ac.in} \\
	\And
    {\includegraphics[scale=0.06]{orcid.pdf}\hspace{1mm}Sayali Mate} \\
	Department of Computer Engineering\\
	COEP Technological University\\
	Pune, India 411005 \\
	\texttt{matesa20.comp@coeptech.ac.in} \\
 \And
    {\includegraphics[scale=0.06]{orcid.pdf}\hspace{1mm}Chinmayee Kulkarni} \\
	Department of Computer Engineering\\
	COEP Technological University\\
	Pune, India 411005 \\
	\texttt{kulkarnicn20.comp@coeptech.ac.in} \\
 \And
    {\includegraphics[scale=0.06]{orcid.pdf}\hspace{1mm}Prof. Suraj Sawant} \\
	Department of Computer Engineering\\
	COEP Technological University\\
	Pune, India 411005 \\
	\texttt{sts.comp@coeptech.ac.in} \\
}

\renewcommand{\shorttitle}{\textit{arXiv} Template}

\hypersetup{
pdftitle={A template for the arxiv style},
pdfsubject={q-bio.NC, q-bio.QM},
pdfauthor={David S.~Hippocampus, Elias D.~Striatum},
pdfkeywords={First keyword, Second keyword, More},
}

\begin{document}
\maketitle

\begin{abstract}
Air pollution is a significant health concern worldwide, contributing to various respiratory diseases. Advances in air quality mapping, driven by the emergence of smart cities and the proliferation of Internet-of-Things sensor devices, have led to an increase in available data, fueling momentum in air pollution forecasting.
The objective of this study is to devise an integrated approach for predicting air quality using image data and subsequently assessing lung disease severity based on Air Quality Index (AQI).The aim is to implement an integrated approach by refining existing techniques to improve accuracy in predicting AQI and lung disease severity.The study aims to forecast additional atmospheric pollutants like AQI, PM10, O3, CO, SO2, NO2 in addition to PM2.5 levels.Additionally, the study aims to compare the proposed approach with existing methods to show its effectiveness.
The approach used in this paper uses VGG16 model for feature extraction in images and neural network for predicting AQI.In predicting lung disease severity, Support Vector Classifier (SVC) and K-Nearest Neighbors (KNN) algorithms are utilized.
The neural network model for predicting AQI achieved training accuracy of 88.54 \% and testing accuracy of 87.44\%,which was measured using loss function,while the KNN model used for predicting lung disease severity achieved training accuracy of 98.4\% and testing accuracy of 97.5\%
In conclusion, the integrated approach presented in this study forecasts air quality and evaluates lung disease severity, achieving high testing accuracies of 87.44\% for AQI and 97.5\% for lung disease severity using neural network, KNN, and SVC models.
The future scope involves implementing transfer learning and advanced deep learning modules to enhance prediction capabilities. While the current study focuses on India, the objective is to expand its scope to encompass global coverage.
\end{abstract}


\section{Introduction}\label{sec1}
\begin{figure}
    \centering
    \textbf{Fig. 1}\par\medskip
    \includegraphics[width=0.75\linewidth]{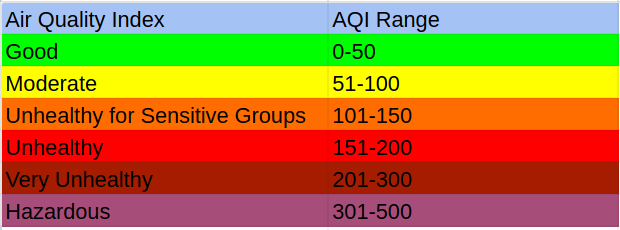}
    \caption{AQI Ranges \cite{'USA-Info'}}
    \label{AQI ranges}
\end{figure}

 Air pollution encompasses any substance or particle that contaminates the air, leading to a decline in air quality and posing health risks. Poor air quality has detrimental effects on various bodily systems, such as the respiratory, cardiovascular, and neurological systems. Prolonged exposure to air pollution can result in the development of respiratory ailments and other lung-related disorders.\cite{12345}. Forecasting AQI reliably and accurately is of utmost importance for ecological preservation and ensuring public health safety. Thus determining the quality of air accurately has become a need of an hour. This information empowers individuals to take actions to reduce their exposure to air pollution and advocate for policies that promote cleaner air and thus in turn helps to live a healthier life.The depicted ranges and interpretations of each level of the Air Quality Index (AQI), as outlined by the Environmental Protection Agency (EPA) and accessible via the AIRNow website, are illustrated in Figure \ref{AQI ranges}.

Amongst 118 countries, India stands 4th \cite{23455} in terms of severe air pollution. Primary pollutants include:PM2.5,PM10,O3, SO2, CO and NO2. Average United States AQI of India was found to be 151 which is very high and thus a matter of concern\cite{23455}. People living in such areas where concentration of the pollutants is high are susceptible respiratory diseases.\cite{Doiron1802140} The United States Environmental Protection Agency has divided the AQI into normalized ranges from 0 to 500 as shown in Figure 1. \cite{'USA-Info'}

 Therefore, as evidenced by the above statistics, the correlation between air quality and lung diseases is profound. Consequently, there is an intention to develop an integrated framework for predicting the AQI from image data. Subsequently, utilizing AQI predictions to forecast the incidence and severity of lung diseases within particular regions.


A comparative analysis of different Regression models has been done by Li, Chenchen et al. 2021\cite{9708644}. Out of all the models, Random Forest Regression (RFR) algorithm and Gradient Boosting Regression (GBR) algorithm exhibited greater suitability for air quality prediction.

In the study done by Fabian Surya Pramudya et al. 2023\cite{10428349} important attributes like SO2,PM2.5,PM10,CO,O3 and NO2 are extracted.Then instead of using ML model a formula is used to calculate AQI for each pollutant.The result AQI is determined by selecting the highest value among all calculated AQI values for respective pollutants. ArcGIS is used to visualise the air pollution.

An analysis of all the algorithms like  random forest, support vector machine, decision tree, and logistic regression algorithms has been done by Agarwal et al. 2022\cite{9965052} for predicting lung cancer.The RF algorithm gives highest accuracy of 92.3 \% among all the algorithms used.

The paper by authors V, Mercy Rajaselvi et al. 2022 \cite{9823787} explores various machine learning models for diagnosing lung diseases, finding promising accuracies of 80\% to 85\%. Notably, Convolutional Neural Networks (CNNs) and Transfer Learning emerge as key methodologies in this domain.

Most of the papers discussed above have confined their predictions limited to a particular area along with less accuracy. Currently, for predicting AQI, only use of PM2.5 is done as it is the primary pollutant, but other factors also play an important role in determining the air quality. Also, till now, predictions for air quality in metropolitan cities for 
a particular country has not been done in the field of air quality detection. Additionally, an integrated approach relating the AQI and the severity of lung disease in a particular area based on personal details is not done till date.

To address the key issues mentioned above, an approach has been developed accordingly. Metropolitan cities in India such as Tamil Nadu, Mumbai, Knowledge Park in Greater Noida, New Industrial Town in Faridabad, ITO in Delhi, Bengaluru, and Dimapur in Nagaland have been selected. Additionally, the approach involves predicting the levels of pollutants such as "AQI," "PM2.5," "PM10," "O3," "CO," "SO2," and "NO2." Furthermore, the severity of lung diseases that individuals living in specific areas may experience is also being predicted based on the pollutant levels.
The proposed approach combines predicting the air quality in a specific city with personalized factors to forecast the severity of lung diseases. Additionally, a customized neural network model has been developed to achieve higher accuracy compared to other CNN models.

Section~\ref{sec:literature-review} covers literature review,Section~\ref{sec:methodology} explains methodology,Section~\ref{sec:results} discusses the implementation and results,Section~\ref{sec:discussion} talks about discussion and Section~\ref{sec:conclusion} includes conclusion of the research.

\section{Literature Review}\label{sec:literature-review}
\subsection{Air Quality Index}

The study done by Fabian Surya Pramudya et al. 2023 \cite{10428349} investigates air pollution in Jakarta, Indonesia, utilizing data from the Jakarta Pollution dataset and Sentinel-5P satellite imagery. By analyzing six major pollutants and employing GIS software for visualization, the study determines Jakarta's AQI.

The aforementioned study primarily focuses on the visualization and computation of Air Quality Index (AQI) using established pollutant formulas.However, for future insights into AQI trends, subsequent researchers utilized Machine Learning methodologies.


Several studies have explored regression models for air quality estimation. Li, Chenchen et al. 2021 \cite{9708644} investigated the use of Random Forest Regression and Gradient Boosting Regression for determining AQI in Henan Province, China. Their findings suggest that these tree-based regression models outperform other methods in this context.In contrast, Nandini et al. 2019 \cite{9063845}  employed Multinomial Logistic Regression for air quality prediction. Their approach involved data segmentation using K-means clustering, followed by applying Multinomial Logistic Regression to predict pollution levels within each cluster (high, medium, and low). Their study suggests that Multinomial Logistic Regression achieved high accuracy in predicting air quality categories.

The exploration of regression models provides a foundational understanding of traditional statistical approaches, paving the way for a deeper examination of neural network models in subsequent analysis.
The authors Wang Zhenghua et al. 2017 \cite{8446883} presented an improved BP neural network method for predicting AQI. Dataset comprises of daily weather forecast and historical AQI data from two days prior. A genetic algorithm is integrated to enhance the performance. This model has an accuracy of 82.5\%. But as it relies on meteorological data this can limit its predictive accuracy.
The authors Quinli WU and Huaxing Lin, 2019 further try to explore this relationship by proposing a hybrid model that combines several techniques to improve forecasting accuracy in \cite{WU2019808}. The dataset comprises of real-world data from Beijing and Guiliy. A hybrid model uses secondary decomposition (SD), AI method and optimization algorithm along with LSTM(long short-term memory) neural network.


Researchers examined air quality in distinct locations using various techniques.Liu et al. 2019 \cite{app9194069}  concentrated on Beijing, China, where they predicted AQI values while prioritizing NO$_2$. This approach involved analyzing datasets from Beijing containing hourly measurements of common air pollutants.In contrast, Kumar et al.2021 \cite{9675669}  investigated Bengaluru, India. Their study spanned six years (2015-2020) and utilized data collected from ten separate stations within the city. They employed a combination of data analysis methods, including selection of key pollutants, clustering (EM), and classification (J48) to assess air quality and visualize AQI levels.

\subsection{Lung Disease Severity}

The research by authors Dany Doiron et al. 2019\cite{Doiron1802140} and Gabriel-Petrică Bălă et.al, 2021\cite{Bălă2021} explores the mappings between air pollution and its effect on different lung diseases and shows that greater exposure of the pollutants to more is the severity of the disease. Additional research has been conducted in the papers listed below, utilizing various machine learning models to explore these relationships.

The authors Yan Jin et.al, 2023\cite{10.3389/feart.2023.1105140} have used linear as well as non-linear models for exploring the impacts of air pollution on the respiratory diseases. The findings indicate that the Random Forest model outperformed others, suggesting that non-linear models generally yield superior performance.

The study by Agarwal, Shubhada et al. 2022\cite{9965052} uses ML models like support vector machine,random forest,logistic regression and decision tree algorithms to predict lung cancer based on a dataset consisting of 13 parameters.The RF algorithm gives highest accuracy of 92.3 \% among all the algorithms used.

In the paper by authors V, Mercy Rajaselvi et al. 2022\cite{9823787}, an examination is conducted on various machine learning models utilized in diagnosing lung diseases. The findings reveal that numerous studies have been done into the application of ML techniques for diagnosing a spectrum of lung ailments including lung cancer, COVID-19, and pneumonia. These investigations consistently report promising accuracies ranging from 80\% to 85\%. Particularly, Convolutional Neural Networks (CNNs) and Transfer Learning stand out as predominant methodologies in this field.


\section{Methodology}\label{sec:methodology}
The methodology outlined in this study has been formulated through a detailed analysis of the research gaps discussed earlier. By examining existing literature, areas where prior studies lacked sufficient coverage were identified. The methodology presented aims to address these gaps by employing techniques and approaches as discussed below.
\begin{figure}
    \centering
    \textbf{Fig. 2}\par\medskip
    \includegraphics[width=0.90\linewidth]{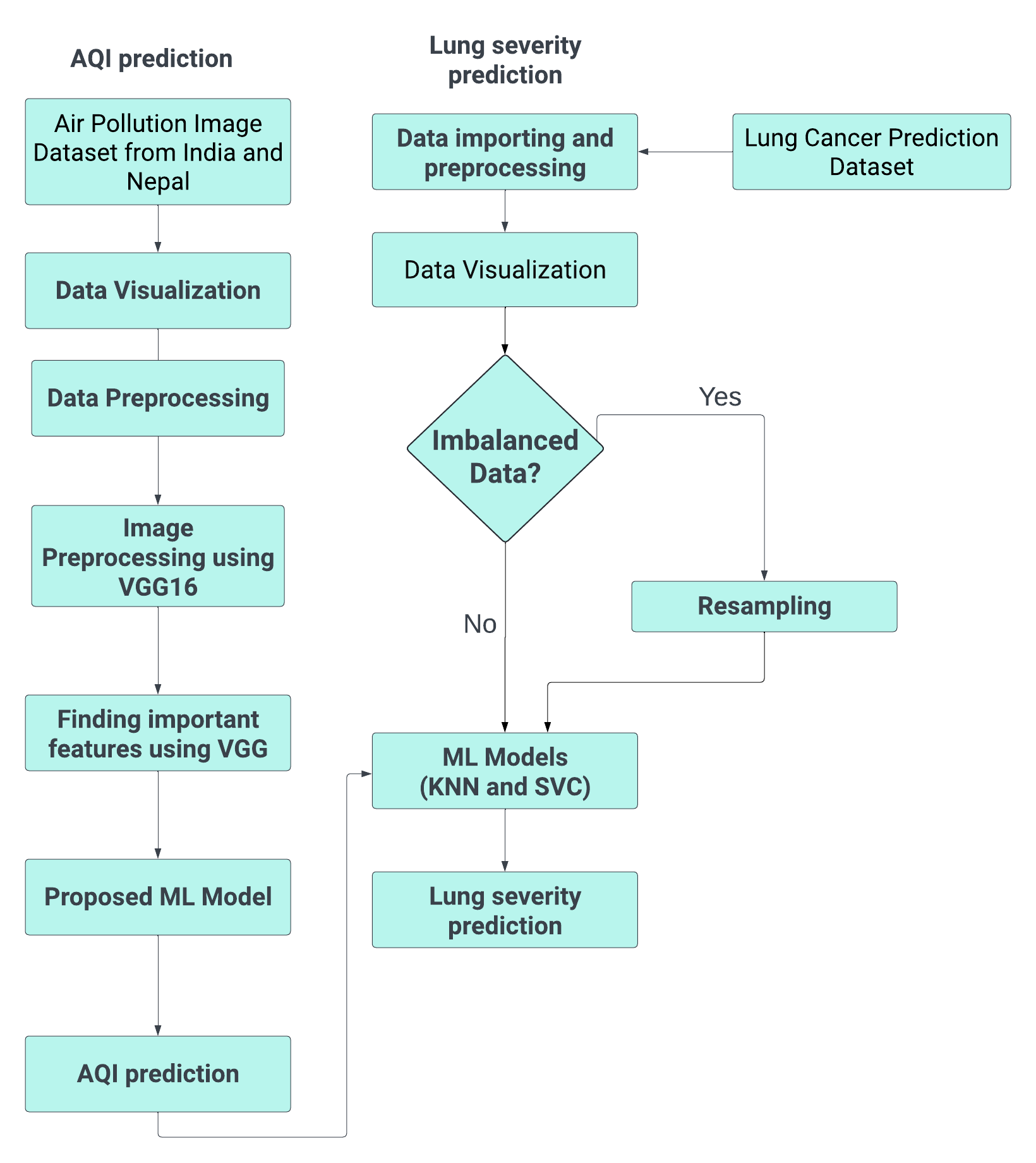}
    \caption{Design Model}
    \label{fig:enter-label}
\end{figure}
\newpage
\subsection{Prediction of AQI from image dataset}
\begin{itemize}
\item Dataset: The dataset utilized for predicting various pollutants is the 'Air Pollution Image Dataset from India and Nepal,' which is publicly accessible on Kaggle. \cite{adarsh_rouniyar_sapdo_utomo_john_a_pao-ann_hsiung_2023}.
\item Data Preprocessing: The initial dataset comprised 12,240 entries, including images from countries such as India and Nepal. To narrow the scope of the project to India exclusively, only images of Indian cities are being extracted.Furthermore, to enhance result accuracy, any null values within the dataset were removed during the preprocessing stage.
\item Data Visualization
\begin{figure}[htbp]
  \centering
  \begin{minipage}{0.5\textwidth}
    \centering
    \textbf{Fig. 3}\par\medskip
    \includegraphics[width=\linewidth]{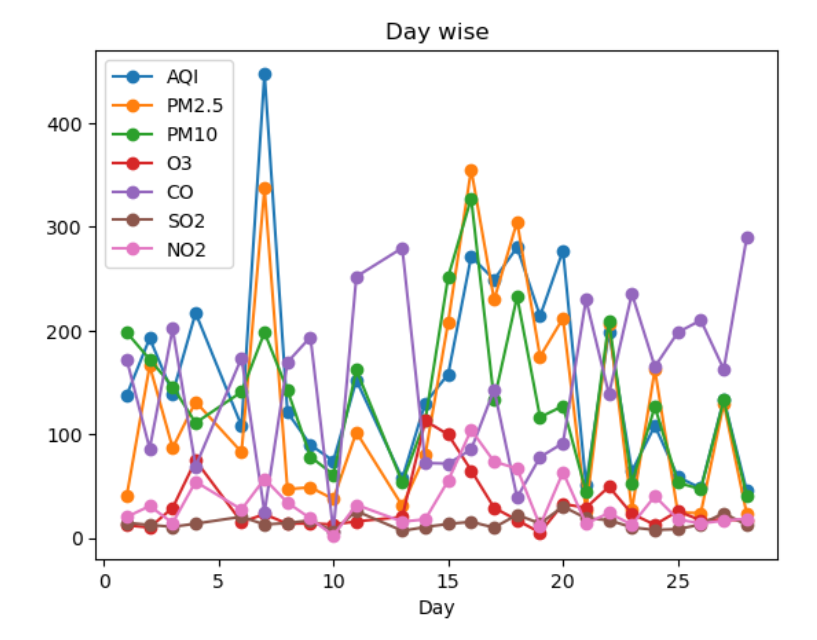}
    \caption{Day-wise Distribution}
    \label{fig:day-wise}
  \end{minipage}\hfill
  \begin{minipage}{0.5\textwidth}
    \centering
    \textbf{Fig. 4}\par\medskip
    \includegraphics[width=\linewidth]{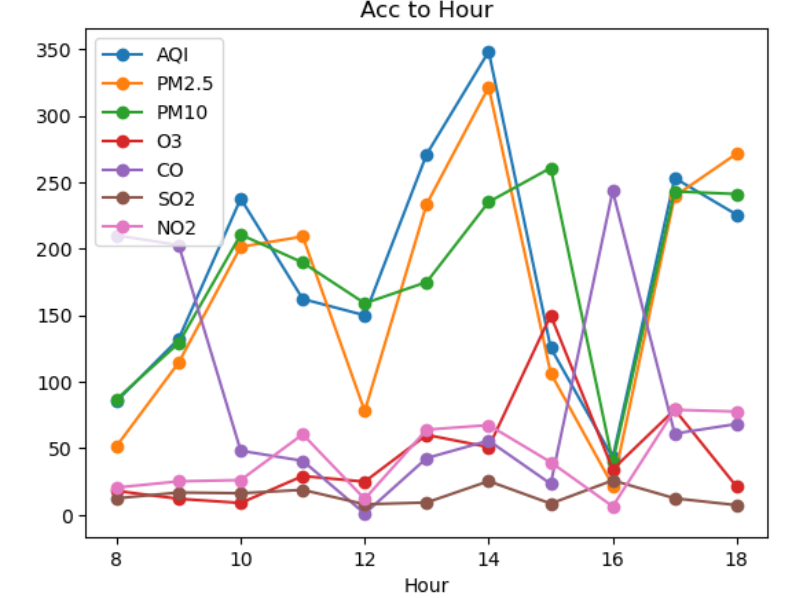}
    \caption{Hour-wise Distribution}
    \label{fig:hour-wise}
  \end{minipage}\hfill
\end{figure}
\\
 \begin{figure}
    \begin{minipage}{1\textwidth}
    \centering
    \textbf{Fig. 5}\par\medskip
    \includegraphics[width=\linewidth]{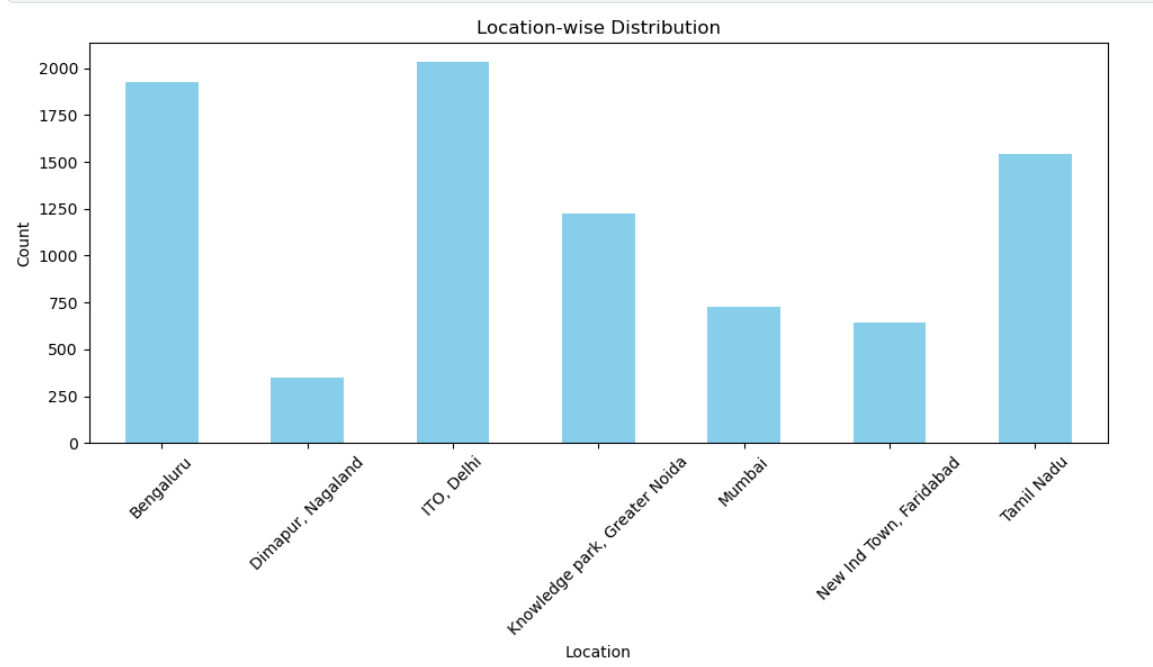}
    \caption{Location wise Distribution}
    \label{fig:location-wise}
    \end{minipage}
    \end{figure}
\item Image Preprocessing: The image was loaded and converted into a numpy array, storing the values of each pixel. This representation converts the image into numerical form. Further preprocessing operations such as mean subtraction and scaling to pixel values were conducted on the image to prepare it for the VGG16 model. Subsequently, the inbuilt VGG16 function with pre-trained ImageNet weights was utilized for image preprocessing and feature extraction.
    \item Feature Extraction: Once the pre-trained model is loaded,its layers are freezed to ensure that is doesn't update the weights of pre-trained VGG layers during training.Next, a Global Average Pooling layer is incorporated atop the VGG16 model's output, serving to compress spatial dimensions and generate a consistent-length feature vector. After adding the layer,preprocessed image is passed through the feature extractor to obtain a feature vector.After that ,these features are combined with Metadata.
    \item Models used -
    Further, a custom neural network model was built and used for AQI prediction. The model architecture is shown in the figure below:-
    \begin{figure}
        \centering
        \textbf{Fig. 6}\par\medskip
        \includegraphics[width=1\linewidth]{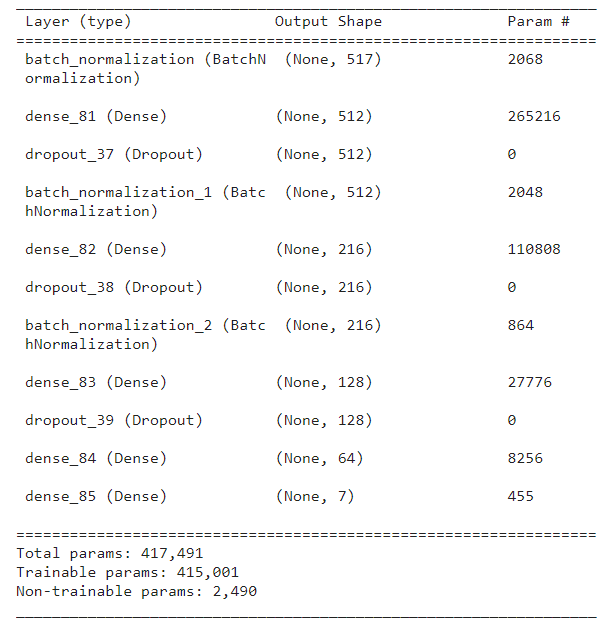}
        \caption{AQI Prediction Model Architecture}
        \label{fig:enter-label}
    \end{figure}
    The results obtained from this model were then used to predict the severity of the lung diseases. 
\end{itemize}
\subsection{Prediction of lung disease from AQI}
\begin{itemize}
\item Dataset: This dataset is taken from kaggle\cite{dataset2}.The dataset consists of 1000 instances.
\item Data Understanding and Preprocessing:In the initial dataset, there were 25 features. However, after careful consideration, the selection was narrowed down to the 11 most relevant features. The severity of lung disease is assessed on a scale ranging from 1 to 7. During data preprocessing, an imbalance was observed in the representation of the output feature, "chronic lung disease." To address this, resampling was performed, resulting in the dataset instances growing to 1550.
\item Data visualization:
 \begin{figure}[!h]
    \centering
    \textbf{Fig. 7}\par\medskip
        \includegraphics[width=\textwidth]{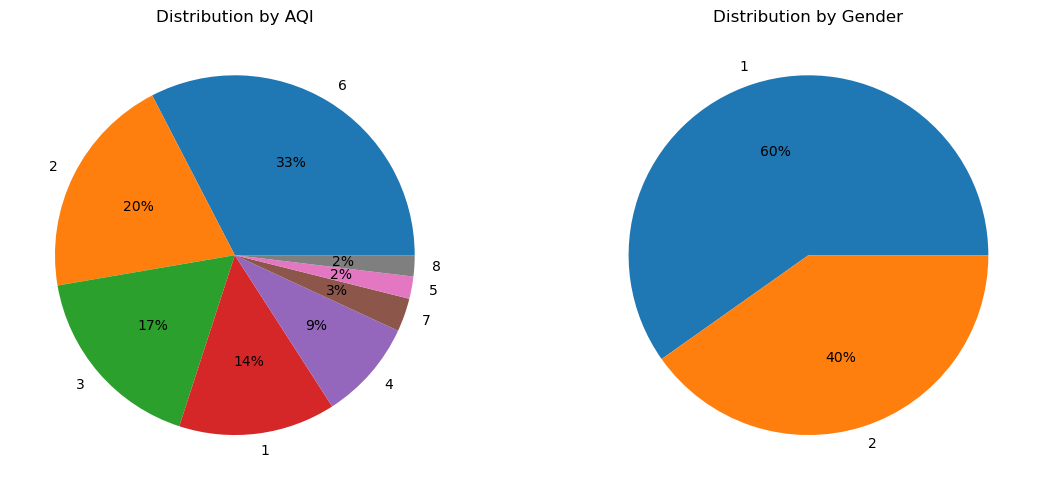}
        \caption{Distribution of dataset by Age and Gender}
        \label{fig:img9}
    \hfill
    \end{figure}
\item Resampling : Resampling techniques were used to balance the imbalanced dataset.SVMSMOTE was utilized for oversampling and RandomUnderSampler for undersampling, combined within an imbalanced-learn Pipeline. 
\item Models used:
Two distinct models, K-Nearest Neighbors (KNN) and Support Vector Classifier (SVC), were utilized for predicting lung disease severity across a scale from 1 to 7,1 being least severe and 7 being the most. KNN operates by assigning a class label to an input based on the majority class among its k nearest neighbors. Conversely, Support Vector Classification (SVC) constructs a hyperplane or a series of hyperplanes within a high-dimensional space to effectively differentiate between instances of distinct classes. Both models were selected for their efficacy in classification tasks and their potential to provide accurate predictions regarding the severity of lung diseases.

\end{itemize}
\section{Implementation and Results}\label{sec:results}
\subsection{Specifications}

The research utilized an NVIDIA GP100 GPU (Pascal architecture, 3584 CUDA cores, 16GB HBM2 memory) for its high performance in training and inference workloads. The software environment consisted of Jupyter Notebook as the development platform and Python as the primary language. Libraries including NumPy, Pandas, Matplotlib, TensorFlow, and scikit-learn provided functionalities for data manipulation, visualization, and machine learning tasks.
\subsection{Results}
    1. AQI Prediction: For Air Quality Prediction, the VGG16 model was utilized for preprocessing, followed by a customized neural network model for predicting the Air Quality class based on the AQI values obtained. The model achieved a training accuracy of 88.54\%, while the testing accuracy reached 87.44\%. The graph below illustrates the loss incurred during the model training. Please refer to Figure 5 for the visualization of the validation and training graph.
    \begin{figure*}[!h]
  \centering
  \begin{minipage}{0.8\textwidth}
    \centering
    \textbf{Fig. 8}\par\medskip
    \includegraphics[width=\linewidth]{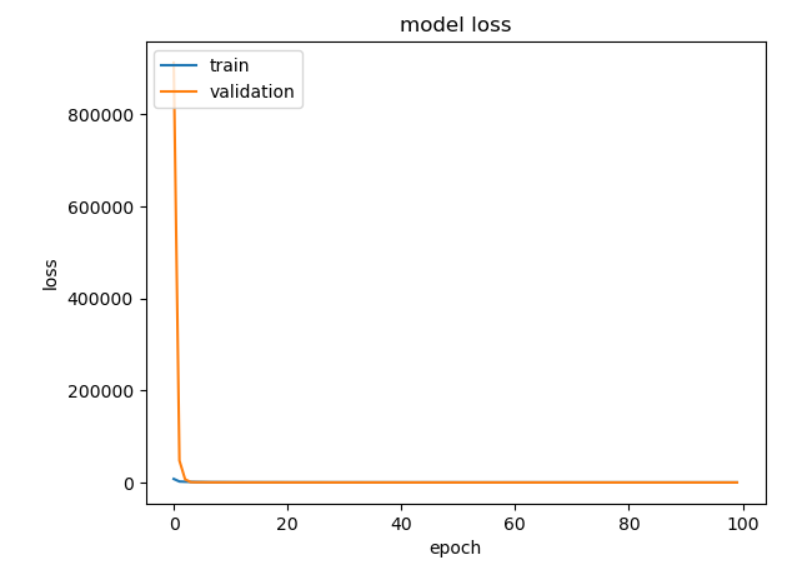}
    \caption{Loss while training the model}
    \label{fig:loss}
  \end{minipage}\hfill
\end{figure*}
\\


2. Models used: for lung disease severity prediction. KNN and SVC models are used whose accuracy is shown in the table below.
\begin{table}[!h]
   \centering
   \caption{Models and Accuracy}
   \begin{tabular}{|c|c|c|} 
       \hline
       \textbf{Model} & \textbf{Training accuracy} & \textbf{Testing accuracy} \\
       \hline
       KNN &  98.4\% & 97.5\% \\
        \hline
        SVC & 79.6\% & 77.0\% \\
        \hline
   \end{tabular}
\end{table}
\\
    3. Frontend dashboard created using streamlit which is a python library. Shown in figures 6 and 7.
\begin{figure*}[!h]
  \centering
  \begin{minipage}{0.45\textwidth}
    \centering
    \textbf{Fig. 9}\par\medskip
    \includegraphics[width=\linewidth]{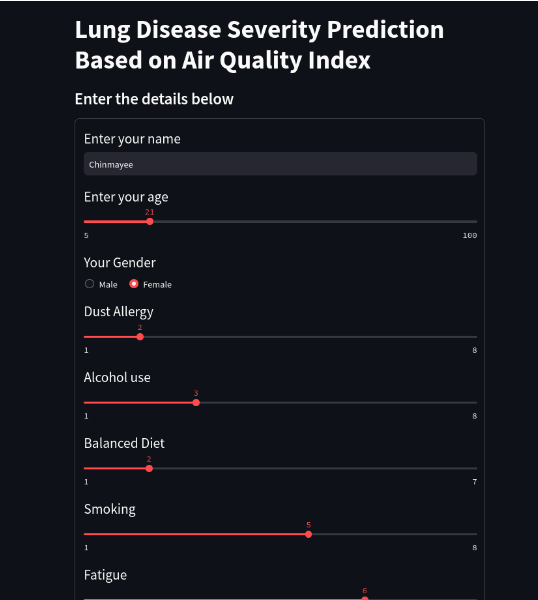}
    \caption{User Input}
    \label{fig:day-wise}
  \end{minipage}\hfill
  \begin{minipage}{0.45\textwidth}
    \centering
    \textbf{Fig. 10}\par\medskip
    \includegraphics[width=\linewidth]{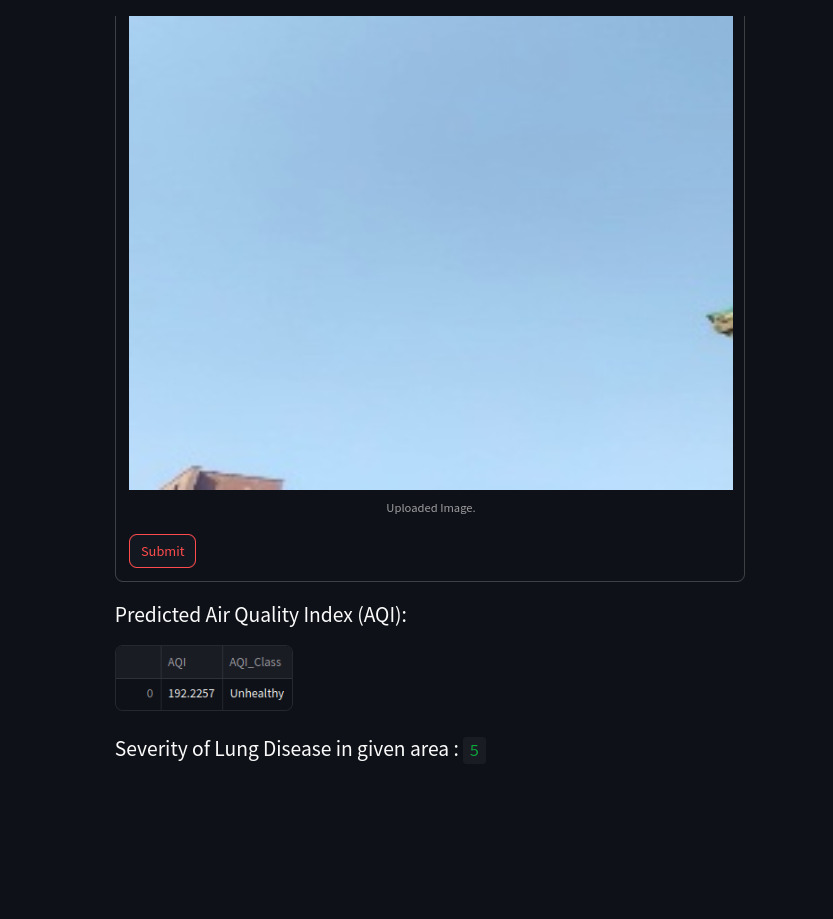}
    \caption{AQI and Lung disease severity predictions}
    \label{fig:hour-wise}
  \end{minipage}\hfill
\end{figure*}

\newpage
\newpage
\section{Discussion}\label{sec:discussion}
The paper employs approaches designed to address existing research gaps. While CNN models are prevalent in this field, the accuracies reported in the literature review typically hover around 80\%. To improve accuracy, a customized convolutional neural network (CNN) model was developed, outlined in Fig. 6. The model comprises 5 dense layers, with batch normalization and dropout layers incorporated in between to mitigate overfitting. 
The authors Yan Jin et.al, 2023\cite{10.3389/feart.2023.1105140} had suggested that for determining the lung disease severity from air pollution, non-linear models perform better than linear models. To validate this assertion, two non-linear models, namely K-Nearest Neighbors (KNN) and Support Vector Classifier (SVC), were utilized. KNN demonstrated a training accuracy of 98.4\% and a testing accuracy of 97.5\%, whereas SVC exhibited a training accuracy of 79.4\% and a testing accuracy of 77.0\%.


\section{Conclusion}\label{sec:conclusion}
Air pollution is one of the major factors for health concern worldwide, contributing to various respiratory diseases. Understanding the severity and early detection can help in prevention of respiration illnesses and thereby enhance common population health.

An integrated approach has been developed to simultaneously forecast air quality and assess the occurrence and severity of lung diseases. This system achieves its objectives by employing novel algorithms to predict multiple pollutants, such as AQI, PM10, O3, CO, SO2, and NO2, alongside PM2.5.In addition to achieving the stated objectives, this study provides comparative analyses with similar models and approaches.

For AQI prediction, a customized neural network model is used.It achieves training accuracy of 88.54\% whereas testing accuracy 87.44\%. Apart from this KNN and SVC models are used for predicting lung disease severity with testing accuracy of 97.5\% and 77.0\% respectively.

A comparative evaluation of proposed neural network model in relation to the work done by authors Sami Kabir et al. 2022 \cite{KABIR2022117905} is provided in the table below:
\begin{table}[htbp]
   \centering
     \caption{Comparative Analysis}
     \begin{tabular}{|c|c|c|} 
         \hline
         \textbf{Model/Paper} & \textbf{Testing MSE} & \textbf{$R^2$}\\
         \hline
        Proposed Model & 96.64 & 0.97\\
          \hline
          \cite{KABIR2022117905} &  1372 & 0.42\\
          \hline
         
     \end{tabular}
\end{table}
\\

While the current scope is limited to India, the objective is to expand its scope to encompass global coverage. Further, the future scope involves advanced deep learning modules and implementing transfer learning to further improve the accuracy of its prediction. Additionally, to strengthen the system's resilience, data harmonization could be implemented by training the model with diverse image formats. This approach would ensure consistency across different data types, increasing the dataset's versatility and enhancing the model's capacity to handle various image formats effectively.

\section{Data Availability}
The dataset used in this study for predicting air quality index is publicly available on Kaggle \url{https://www.kaggle.com/ds/3152196}. Researchers interested in accessing the dataset can obtain it directly from the Kaggle platform.

The dataset used in this study for predicting lung disease severity is publicly available on Kaggle \url{https://www.kaggle.com/datasets/thedevastator/cancer-patients-and-air-pollution-a-new-link/data}. Researchers interested in accessing the dataset can find it on the Kaggle platform.

\bibliographystyle{plainnat}

\begin{thebibliography}{1}

	\bibitem{WU2019808}
	Qunli Wu and Huaxing Lin.
	\newblock A novel optimal-hybrid model for daily air quality index prediction considering air pollutant factors.
	\newblock In {\em Science of The Total Environment}, volume 683, pages 808-821. Elsevier, 2019.
	\newblock ISSN: 0048-9697.
	\newblock DOI: 10.1016/j.scitotenv.2019.05.288.
	\newblock URL: https://www.sciencedirect.com/science/article/pii/S0048969719323290.
	\newblock Keywords: Air quality index (AQI) forecasting, Secondary decomposition (SD), Long short-term memory (LSTM) neural network, Least squares support vector machine (LSSVM), Air pollutant.

\bibitem{8446883}
	Zhenghua Wang and Zhihui Tian.
	\newblock Prediction of Air Quality Index Based on Improved Neural Network.
	\newblock In {\em 2017 International Conference on Computer Systems, Electronics and Control (ICCSEC)}, pages 200-204, 2017.
	\newblock DOI: 10.1109/ICCSEC.2017.8446883.

\bibitem{9965052}
	Shubhada Agarwal, Sanjeev Thakur, and Alka Chaudhary.
	\newblock Prediction of Lung Cancer Using Machine Learning Techniques and their Comparative Analysis.
	\newblock In {\em 2022 10th International Conference on Reliability, Infocom Technologies and Optimization (Trends and Future Directions) (ICRITO)}, pages 1-5, 2022.
	\newblock DOI: 10.1109/ICRITO56286.2022.9965052.

\bibitem{12345}
	Air pollution effects on your lungs, including lung cancer.
	\newblock {\em https://www.asthmaandlung.org.uk/living-with/air-pollution/your-lungs:~:text=Being

\bibitem{23455}
	IQAir.
	\newblock Live most polluted major city ranking.
	\newblock {\em https://www.iqair.com/in-en/world-air-quality-ranking}.

\bibitem{'USA-Info'}
	United States government.
	\newblock Patient Exposure and the Air Quality Index.
	\newblock {\em https://www.epa.gov/pmcourse/patient-exposure-and-air-quality-index}.

\bibitem{adarsh_rouniyar_sapdo_utomo_john_a_pao-ann_hsiung_2023}
	Adarsh Rouniyar, Sapdo Utomo, John A, and Pao-Ann Hsiung.
	\newblock Air Pollution Image Dataset from India and Nepal.
	\newblock Publisher: Kaggle, 2023.
	\newblock URL: https://www.kaggle.com/ds/3152196.
	\newblock DOI: 10.34740/KAGGLE/DS/3152196.

\bibitem{dataset2}
	\newblock https://www.kaggle.com/datasets/thedevastator/cancer-patients-and-air-pollution-a-new-link/data.

\bibitem{9708644}
	Chenchen Li, Yan Li, and Yubin Bao.
	\newblock Research on Air Quality Prediction Based on Machine Learning.
	\newblock In {\em 2021 2nd International Conference on Intelligent Computing and Human-Computer Interaction (ICHCI)}, pages 77-81, 2021.
	\newblock DOI: 10.1109/ICHCI54629.2021.00022.

\bibitem{Doiron1802140}
	Dany Doiron, Kees de Hoogh, Nicole Probst-Hensch, Isabel Fortier, Yutong Cai, Sara De Matteis, and Anna L. Hansell.
	\newblock Air pollution, lung function and COPD: results from the population-based UK Biobank study.
	\newblock {\em European Respiratory Journal}, volume 54, number 1, page 1802140, 2019.
	\newblock DOI: 10.1183/13993003.02140-2018.
	\newblock URL: https://erj.ersjournals.com/content/54/1/1802140.

\bibitem{10.3389/feart.2023.1105140}
	Yan Ji, Xiefei Zhi, Ying Wu, Yanqiu Zhang, Yitong Yang, Ting Peng, and Luying Ji.
	\newblock Regression analysis of air pollution and pediatric respiratory diseases based on interpretable machine learning.
	\newblock {\em Frontiers in Earth Science}, volume 11, page 1105140, 2023.
	\newblock DOI: 10.3389/feart.2023.1105140.
	\newblock ISSN: 2296-6463.
	\newblock URL: https://www.frontiersin.org/articles/10.3389/feart.2023.1105140.

\bibitem{Singh2023}
	Tamanpreet Singh, Amandeep Kaur, Sharon Kaur Katyal, Simran Kaur Walia, Geetika Dhand, Kavita Sheoran, Sachi Nandan Mohanty, M. Ijaz Khan, Fuad A. Awwad, and Emad A. A. Ismail.
	\newblock Exploring the relationship between air quality index and lung cancer mortality in India: predictive modeling and impact assessment.
	\newblock {\em Scientific Reports}, volume 13, number 1, page 20256, November 20, 2023.
	\newblock DOI: 10.1038/s41598-023-47705-5.
	\newblock ISSN: 2045-2322.
	\newblock URL: https://doi.org/10.1038/s41598-023-47705-5.

\bibitem{Bălă2021}
	Gabriel-Petrică Bălă, Ruxandra-Mioara Râjnoveanu, Emanuela Tudorache, Radu Motișan, and Cristian Oancea.
	\newblock Air pollution exposure---the (in)visible risk factor for respiratory diseases.
	\newblock {\em Environmental Science and Pollution Research}, volume 28, number 16, pages 19615-19628, April 1, 2021.
	\newblock DOI: 10.1007/s11356-021-13208-x.
	\newblock ISSN: 1614-7499.
	\newblock URL: https://doi.org/10.1007/s11356-021-13208-x.

\bibitem{9823787}
	Mercy Rajaselvi V, Sanjith J, Samuel Koshy, and Niranjan G M.
	\newblock A Survey on Lung Disease Diagnosis using Machine Learning Techniques.
	\newblock In {\em 2022 2nd International Conference on Advance Computing and Innovative Technologies in Engineering (ICACITE)}, pages 01-04, 2022.
	\newblock DOI: 10.1109/ICACITE53722.2022.9823787.

\bibitem{10004651}
	Syed Krar Haider Bukhari and Labiba Fahad.
	\newblock Lung Disease Detection using Deep Learning.
	\newblock In {\em 2022 17th International Conference on Emerging Technologies (ICET)}, pages 154-159, 2022.
	\newblock DOI: 10.1109/ICET56601.2022.10004651.

\bibitem{10428349}
	Fabian Surya Pramudya, Felix Indra Kurniadi, and Aldilla Noor Rakhiemah.
	\newblock Breathing in Jakarta: Uncovering the Air Quality Index using Data Visualization.
	\newblock In {\em 2023 7th International Conference on New Media Studies (CONMEDIA)}, pages 162-166, 2023.
	\newblock DOI: 10.1109/CONMEDIA60526.2023.10428349.

\bibitem{9675669}
	R Senthil Kumar, Anidha Arulanandham, and Suresh Arumugam.
	\newblock Air quality index analysis of Bengaluru city air pollutants using Expectation Maximization clustering.
	\newblock In {\em 2021 International Conference on Advancements in Electrical, Electronics, Communication, Computing and Automation (ICAECA)}, pages 1-4, 2021.
	\newblock DOI: 10.1109/ICAECA52838.2021.9675669.

\bibitem{app9194069}
	Huixiang Liu, Qing Li, Dongbing Yu, and Yu Gu.
	\newblock Air Quality Index and Air Pollutant Concentration Prediction Based on Machine Learning Algorithms.
	\newblock In {\em Applied Sciences}, volume 9, number 19, page 4069, 2019.
	\newblock DOI: 10.3390/app9194069.
	\newblock URL: https://www.mdpi.com/2076-3417/9/19/4069.

\bibitem{9063845}
	K. Nandini and G. Fathima.
	\newblock Urban Air Quality Analysis and Prediction Using Machine Learning.
	\newblock In {\em 2019 1st International Conference on Advanced Technologies in Intelligent Control, Environment, Computing \& Communication Engineering (ICATIECE)}, pages 98-102, 2019.
	\newblock DOI: 10.1109/ICATIECE45860.2019.9063845.

\bibitem{KABIR2022117905}
	Sami Kabir, Raihan Ul Islam, Mohammad Shahadat Hossain, and Karl Andersson.
	\newblock An integrated approach of Belief Rule Base and Convolutional Neural Network to monitor air quality in Shanghai.
	\newblock In {\em Expert Systems with Applications}, volume 206, page 117905, 2022.
	\newblock DOI: 10.1016/j.eswa.2022.117905.
	\newblock ISSN: 0957-4174.
	\newblock URL: https://www.sciencedirect.com/science/article/pii/S0957417422011514.
}

\end{thebibliography}

\end{document}